\documentclass[10pt,twocolumn,letterpaper]{article}

\usepackage{wacv}
\usepackage{times}
\usepackage{epsfig}
\usepackage{graphicx}
\usepackage{amsmath}
\usepackage{amssymb}
\usepackage{comment}

\usepackage{soul}

\def\wacvPaperID{5} 

\wacvfinalcopy 

\ifwacvfinal
\def\assignedStartPage{1} 
\fi

\ifwacvfinal
\usepackage[breaklinks=true,bookmarks=false]{hyperref}
\else
\usepackage[pagebackref=true,breaklinks=true,colorlinks,bookmarks=false]{hyperref}
\fi

\ifwacvfinal
\else
\fi

\begin{document}

\title{Interpretable Deep Learning-Based Forensic Iris Segmentation and Recognition}

\author{Andrey Kuehlkamp~~~~~~Aidan Boyd~~~~~~Adam Czajka~~~~~~Kevin Bowyer~~~~~~Patrick Flynn\\
University of Notre Dame, Notre Dame, IN, USA\\
{\tt\small \{akuehlka, aboyd3, aczajka, kwb, flynn\}@nd.edu} \\\\
Dennis Chute~~~~~~Eric Benjamin\\
Duchess County Medical Examiner's Office, Poughkeepsie, NY, USA \\
{\tt\small \{dchute, ebenjamin\}duchessny.gov}
}

\maketitle

\begin{abstract}
   Iris recognition of living individuals is a mature biometric modality that has been adopted globally from governmental ID programs, border crossing, voter registration and de-duplication, to unlocking mobile phones. On the other hand, the possibility of recognizing deceased subjects with their iris patterns has emerged recently. 
   In this paper, we present an end-to-end deep learning-based method for postmortem iris segmentation and recognition with a special visualization technique intended to support forensic human examiners in their efforts.
   The proposed postmortem iris segmentation approach outperforms the state of the art and -- in addition to iris annulus, as in case of classical iris segmentation methods -- detects abnormal regions caused by eye decomposition processes, such as furrows or irregular specular highlights present on the drying and wrinkling cornea. The method was trained and validated with data acquired from 171 cadavers, kept in mortuary conditions, and tested on subject-disjoint data acquired from 259 deceased subjects. To our knowledge, this is the largest corpus of data used in postmortem iris recognition research to date. The source code of the proposed method are offered with the paper. The test data will be available through the National Archive of Criminal Justice Data (NACJD) archives.
\end{abstract}

\section{Introduction}

Iris recognition has been a solid biometric identification means for almost three decades \cite{Daugman1993}, observing successful governmental and consumer-level deployments. These include national ID programs such as AADHAAR in India \cite{AADHAAR} (currently the largest biometric identification program worldwide), voter de-duplication in Ghana, Somaliland and Tanzania \cite{elsevierBTT_dedupe}, border control \cite{Daugman2015}, biometric ATMs or unlocking mobile devices \cite{BBVA}. With the FBI's Next Generation Identification system \cite{fbi_2016} gaining momentum, and with recent quest for contactless biometric recognition approaches due to the COVID-19 pandemic, iris recognition is more often brought to the table as a strong candidate for reliable, fast and hygienic human recognition method.

One of the strengths of iris as a biometric identifier is its high temporal stability compared to other biometric modes \cite{irexvi}. Although the assessment of how biological changes impact the systems' decisions (not only similarity scores) remains still one of the research challenges awaiting an authoritative answer \cite{Bowyer_IET_2015,Grother_IET_2015,Ortiz_BTAS_2013,Czajka_BEST_2013,Trokielewicz_BF_2015}, observations from operational deployments suggest that iris recognition may be a good candidate for a ``cradle-to-grave'' biometrics across all human identification approaches. However, one of the recent discoveries made by several research groups was that iris recognition {\it is} possible to be done also after death \cite{Bolme_BTAS_2016,trokielewicz_icb_2016}. This is gruesome and fascinating at the same time. Gruesome, since the possibility of using a deceased person's eyeball for authentication calls for new presentation attack detection mechanisms \cite{trokielewicz_btas_2018}. Fascinating, as it brings another, potentially very fast and reliable element of the forensic toolkit. This paper proposes the first, known to us, fully deep learning-based and open-sourced methodology of processing and matching iris images captured from cadavers in near-infrared light (following the ISO-compliant acquisition protocol \cite{ISO_29794_6}) with the primary goal of providing interpretable visualizations, crucial in deploying iris as an element of forensic examination. 

The method proposed in this paper was designed on the largest, known to us, dataset of iris images captured from 430 deceased subjects. Data from 171 subjects was used for training and validation, and sequestered data from another 259 cadavers was used for testing. Our approach (Fig. \ref{fig:overview})  consists of three elements, which are at the same time the {\bf main contributions of this work}: 

\begin{figure*}[t]
\begin{center}
   \includegraphics[width=1.0\linewidth,trim={0 10 0 0}]{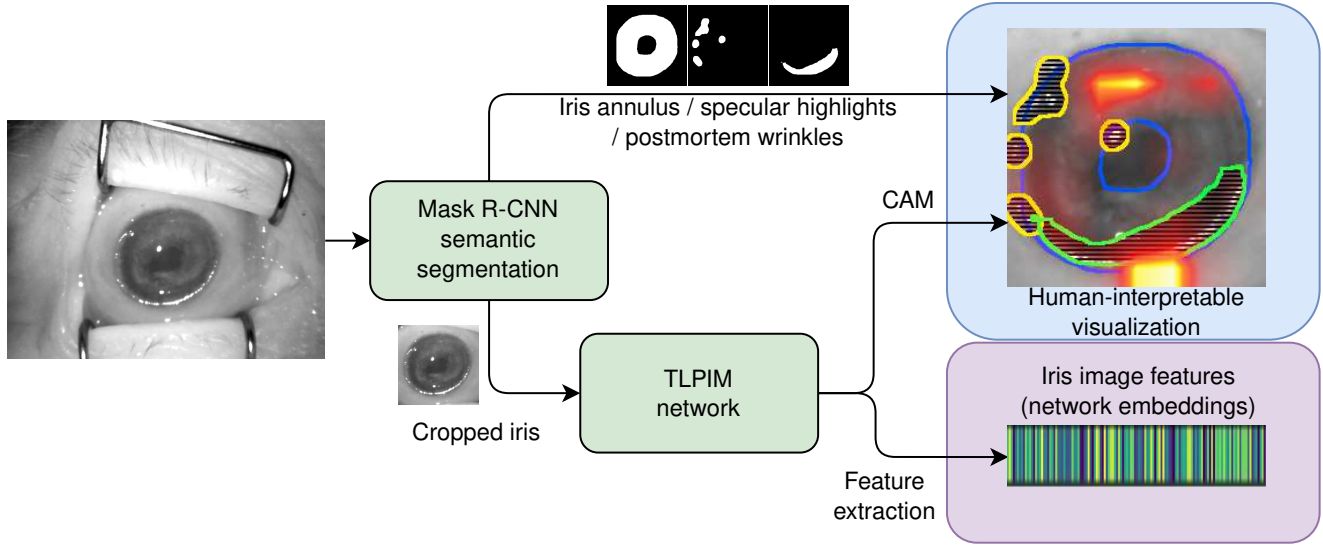}
\end{center}
   \caption{Overview of the proposed approach: postmortem iris images are first processed by fine-tuned Mask R-CNN, to produce a cropped image of the iris and individual masks for the iris and decomposition artifacts. The cropped image is then fed into our TLPIM network, which produces two outputs: a representation of the input image for distance matching, and a Class Activation Map visualization of the network activations. The latter, combined with the segmentation masks, produces a human-interpretable visualization of the matching outcome.}
\label{fig:overview}
\end{figure*}

\begin{enumerate}
    \item[(a)] multi-class segmentation detecting iris annulus, constellations of specular highlights and the cornea/tissue wrinkles, which appear due to the postmortem decomposition processes;
    \item[(b)] ResNet-50-based iris feature extractor, fine-tuned with the goal of providing appropriate network embeddings for postmortem iris images;
    \item[(c)] a Class Activation Map-based visualization explaining the decisions to human forensic examiners, who -- at this point -- are still the only entities  to make authoritative judgments about the matching of biometric samples when acting in the court of law.
\end{enumerate}

Additionally, as a fourth contribution, we offer source codes of the method available with this paper. The test dataset (images from 259 cadavers) will be submitted to the National Archive of Criminal Justice Data (NACJD) archives \cite{NACJD} and can be requested for research purposes directly at the NACJD.

\section{Related works}

\subsection{Forensic Iris Recognition}

While iris recognition for living individuals has been around for three decades, the recognition of deceased individuals using their iris patterns has long been believed to be impossible. The old assumption of iris recognition inventor that ``soon after death, the pupil dilates considerably, and the cornea becomes cloudy'' \cite{Daugman_BBC_2001} propagated to future research papers and left us with conclusions such as ``the iris ... decays only a few minutes after death'' \cite{Szczepanski_CISIM_2014}. The first evidence that perimortem (acquired right before death) and postmortem iris matching is possible has been however demonstrated by Sansola \cite{Sansola_MastersThesis_2015}, who also observed correct matching results for at least 70\% of cases when only postmortem irises were compared (depending on time after death). Trokielewicz \etal \cite{trokielewicz_icb_2016} proposed the first publicly available dataset of near infrared and visible-light postmortem iris images, and estimated that if bodies are kept in temperatures below $42^{\circ}$F ($6^{\circ}$C), the correct matching can be obtained after 27 hours with state-of-the-art iris matchers. Their extended study \cite{Trokielewicz_TIFS_2019} and research by Sauerwein \etal \cite{Sauerwein_JFO_2017} suggest that correct matches may be expected even from three to five weeks after demise in mortuary and winter-time outdoor conditions, respectively. However, Bolme \etal \cite{Bolme_BTAS_2016} reported much shorter time horizons for successful post-portem iris matching when bodies are kept outdoor during summer, due to severe and rapid decomposition processes happening in the eye. These studies suggest that (a) postmortem iris recognition is possible, and (b) its performance highly depends on ambient conditions. Boyd \etal  \cite{Boyd2020} summarized multiple facets of postmortem iris recognition in a comprehensive survey.

The methods specifically designed for cadaver eyes are sparse and yet emerging. Trokielewicz \etal \cite{trokielewicz_ivc_2020} proposed the first deep learning-based segmentation algorithm based on SegNet model for cadaver eyes, combined with classical (Gabor wavelet-based) and newly designed (data-driven \cite{troki_wacv2020}) feature extractors. {\bf Our work is different from these past efforts} in that (a) the segmentation model, based on Mask R-CNN, instead of detecting only iris annulus detects also other areas affected by decomposition processes, such as tissue wrinkles and irregular corneal specular highlights, (b) leverages a convolutional neural network (CNN) to extract postmortem iris features, (c) provides human-interpretable visualizations of the network's output potentially helpful for forensic examiners, and (d) was designed on the largest known to us research dataset of postmortem iris images acquired from 430 deceased subjects.

\subsection{Deep Learning-Based Iris Recognition}

With iris recognition being a popular means of biometric verification for quite some time, it comes to no surprise that several attempts to improve it are based on deep learning methods. Among the more relevant earliest works in this area can be mentioned DeepIris \cite{Liu2016DeepIrisLP} and DeepIrisNet \cite{Gangwar2016DeepIrisNetDI}. 
These approaches proposed their own architectures to solve the iris recognition problem, but were limited by network design, given the lack of large-scale datasets for training a CNN from scratch. An alternative approach was chosen by Minaee \etal \cite{Minaee2016AnES}, which instead proposed the use of off-the-shelf CNNs to extract iris image features. 
A more complete evaluation of known deep network architectures applied to iris recognition is presented in Nguyen \etal \cite{Nguyen2017}. They show that many CNNs trained for general object classification can be well adapted to work with iris images. A similar study later showed that fine-tuning the pre-trained general purpose CNNs can lead to better results than training from scratch \cite{boyd_btas2019}.

The use of CNNs to extract embeddings into a high-dimensional Euclidean space to measure similarity between images has been around for a few years, but one of the most relevant works in biometrics is Schroff \etal \cite{Schroff_2015_CVPR}, where the authors proposed the use of triplets for training a network to learn face representations. The main idea is to measure the distance between embeddings from an \textit{anchor} sample, a \textit{positive} sample which belongs to the same class as the anchor, and a \textit{negative} sample coming from a different class. A loss function was created to minimize the \textit{anchor-positive} distance at the same that it maximizes the \textit{anchor-negative} distance, effectively separating subjects faces in the embedding space.

Although several works have already used triplet networks for iris recognition \cite{Zhao_2017_ICCV,Ahmad2019,Wang2019IrisIS,Yang_2021_WACV}, this work was specifically inspired by Boyd \etal \cite{boyd_triplet}, who compared the efficacy of the original Gabor-based iris features with data-driven features for recognition, using a triplet model similar to Schroff \etal. Their conclusion showed very similar performance between hand-crafted and data-driven features, also confirming the applicability of a triplet model to iris recognition.

\section{Datasets}
\label{sec:datasets}

\subsection{Segmentation dataset} 
Training a model for semantic segmentation requires annotations of the areas to be demarcated. We used a dataset of images collected from several larger datasets by Trokielewicz \etal~\cite{troki_wacv2020}, used to train SegNet to perform iris segmentation. This dataset contains samples from the following public iris segmentation benchmarks: \textit{ND-0405}, \textit{CASIA}, \textit{BATH}, \textit{BioSec}, \textit{UBIRIS}, and \textit{Warsaw-Biobase-Postmortem-Iris v2.0}. The authors provide two types of ground truth annotation along with the images: \textit{coarse}, where iris annulus and major occlusions are approximated by polygons, and \textit{fine}, where smaller occlusion details like eyelashes and specular highlights were also annotated. This dataset contains a total of 7,193 images.

We want to leverage the ability of Mask R-CNN, the segmentation model used in this work, to detect and segment multiple types of objects in a single pass. This allows to create additional masks that can then be merged into a more accurate boundary of the usable iris texture, but also provides richer information to the forensic examiner about the detected non-usable iris regions. Thus, we manually annotated 179 images with wrinkles and 252 images with specular highlights for a subset of images from the \textit{Warsaw-Biobase-Postmortem-Iris v2.0} dataset.

\subsection{Recognition datasets}
The Warsaw BioBase Postmortem Iris v2.0 (\textit{Warsaw v2.0}) \cite{Trokielewicz_TIFS_2019} is one of the first datasets of postmortem iris images made publicly available. It comprises 1,200 near-infrared (NIR) postmortem images of 37 deceased subjects collected in controlled settings. The time between death and image acquisition (postmortem interval -- PMI) ranges between 5 and 800 hours.

\begin{figure*}[htb]
\begin{center}
   \includegraphics[width=.85\linewidth,trim={0 15 0 0}]{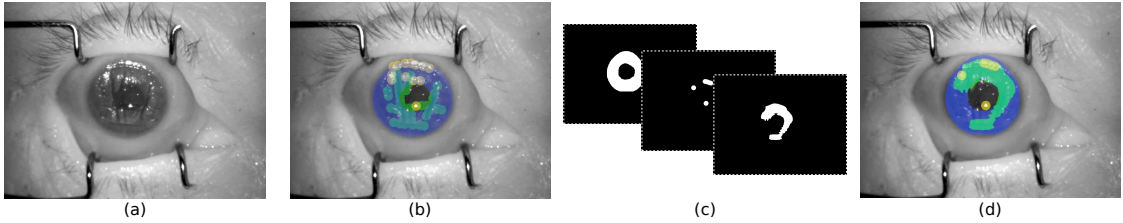}
\end{center}
   \caption{The proposed segmentation of postmortem degraded irises: (a) eyeball presenting wrinkles and irregular specular highlights; (b) ground-truth annotations for iris (blue), wrinkles (green) and highlights (yellow); (c) iris annulus, wrinkles and highlights automatically predicted with Mask R-CNN; (d) overlapping predicted masks, yielding a usable iris area (blue).}
\label{fig:segmentation}
\end{figure*}

Two other datasets used in this work were collected in an operational setting in a medical examiner's office. The first (\textit{DCMEO 01}) contains 621 NIR images from 134 deceased subjects (254 distinct identities) collected in 2019. The images were acquired in up to 9 sessions, where the latest session occurred at most 284 hours after the time of death. The second (\textit{DCMEO 02}) contains 5,770 NIR images from 259 subjects collected in the year of 2020. In this dataset, the longest time elapsed after death was 1,674 hours (~69 days), captured in 53 sessions.

The \textit{DCMEO 02} was kept solely for subject-disjoint testing purposes. A union of \textit{Warsaw v2.0} and \textit{DCMEO 01}, which we call \textit{Combined} dataset, was used in training and validation of the recognition models. These training datasets were collected by two different teams in different mortuary premises, and that makes the training set more diverse.

\section{Multi-object Postmortem Iris Segmentation}
\label{sec:segmentation}

One of the most impactful factors on postmortem iris recognition accuracy is segmentation \cite{Trokielewicz_TIFS_2019}, and the challenge of such images lies in the artifacts that are created by the decomposition process. Dehydration of the eye tissues causes them to shrink and crease. This introduces detrimental features to accurate segmentation such as wrinkles and additional highlights (Fig. \ref{fig:segmentation}a). These are not adequately handled by traditional iris recognition methods, leading to false positive or false negative errors. Our intention is to consider such damaged regions as occlusions and eliminate them from the usable iris area, informing forensic examiners about the potential locations of intact portions of the texture (Fig \ref{fig:segmentation}b-e).

Mask R-CNN \cite{he2017mask} is a network designed for object instance segmentation that is able to not only perform object instance detection, but has also a branch for object mask prediction, and can be adapted to detect a variety of objects. Mask R-CNN was originally trained on the COCO dataset \cite{lin2014microsoft} for general object detection on RGB images. 
We first fine-tuned Mask R-CNN to detect the iris texture in NIR images, using images and labels from a combination of live and postmortem iris images, and then fine-tuned the model to perform instance detection and segmentation using data with newly annotated wrinkles and highlights (as described in Sec. \ref{sec:datasets}). In the first step of Mask R-CNN fine-tuning the parameters used for training were: ResNet-50 backbone with Feature Pyramid Network (FPN), 3x schedule backbone, 8 images per batch, base learning rate 0.00025, 5000 as maximum iterations, 128 proposals per image (\verb!ROI_HEADS.BATCH_SIZE_PER_IMAGE!) and one class (iris). These same settings were employed in further training Mask R-CNN to detect and segment highlights and wrinkles, except for the number of classes increased to 3 (iris, highlights and wrinkles). In both cases, {\it Detectron2} \cite{Detectron2} pre-configured data augmentation settings were used.

\section{Postmortem Iris Recognition}

To create a model capable to perform iris recognition, we chose an implementation of VGGFace2 \cite{cao2018} on the Keras framework, that uses the ResNet-50 backbone \cite{KerasVGG}. This model was first fine-tuned to live iris images, and then trained to generate high dimensional embeddings using triplet loss. Further details of these procedures will be described in this section. We call our model Triplet Loss Postmortem Iris Model (TLPIM).

\begin{figure}[htb]
\begin{center}
   \includegraphics[width=1.0\linewidth,trim={0 100 0 15}]{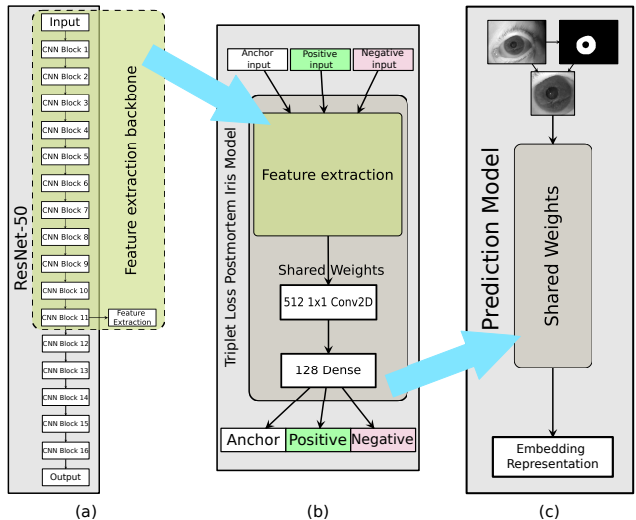}
\end{center}
   \caption{TLPIM Architecture: (a) using feature extraction layers from ResNet-50 VGGFace2 fine-tuned for iris images; (b) composition of a triplet loss model to obtain iris features (network embeddings); (c) the final model for iris encoding extraction.}
\label{fig:architecture}
\end{figure}

\subsection{Tuning VGGFace2 to iris images}
We used CASIA-Iris-Thousand dataset \cite{casiaV4} to fine-tune VGGFace2 to iris images, by training it to perform classification on its 2,000 classes. Images segmented using Mask R-CNN and cropped to a size of $256\times256$ pixels around the detected iris were fed into the network, which was trained for 8 epochs with a learning rate of 0.01 and stochastic gradient descent as the optimizer. After this tuning, features were extracted and classified using an SVM with the RBF kernel, yielding a classification accuracy of 96.28\%, suggesting a satisfactory adaptation to NIR iris images.

\subsection{Adapting VGGFace2 to a triplet loss-based model}
Previous work evaluated the use of ResNet architecture to extract features for iris recognition, and has determined the peak recognition accuracy happens with features extracted at the end of the 11th convolutional block \cite{Nguyen2017}. Following these findings, we extracted features at this layer, and appended a $512\times1\times1$ convolution layer with \textit{Relu} activation for dimensionality reduction (Fig. \ref{fig:architecture}a-b). Using this as a backbone, we build our triplet model network by adding a dense 128-neuron output layer with \textit{sigmoid} activation to ensure the output is contained between 0 and 1. Figure \ref{fig:architecture} illustrates this architecture.

As proposed by Schroff \etal~\cite{Schroff_2015_CVPR}, the triplet loss is expressed by

\begin{equation}
Loss = \sum_{i=1}^{T} [\|f(x_i^a) - f(x_i^p)\|_2^2-\|f(x_i^a) - f(x_i^n)\|^2+\alpha]_+,
\end{equation}

where $T$ is the number of training triplets, $x_i^a, x_i^p, x_i^n$ is an image triplet (anchor, positive and negative), $f$ is the function that extracts embedding representation of images, and $\alpha$ is an enforced margin. One problem with this loss function is that for any difference between positive and negative samples that is smaller than 0, the final result is forced to be 0, eliminating the ability of the network to learn from negative distances. We thus adopted a modification proposed by Arsenault~\cite{arsenault2018} that forces the distance to be positive (through the use of a sigmoid output layer), and adding a logarithmic penalty to the error: 

\begin{equation}
\begin{aligned}
    Loss = {} \sum_{i=1}^{T} &[-\ln(-\frac{(f(x_i^a)-f(x_i^p))^2}{\beta}+1+\epsilon) \\
                &-\ln(\frac{\beta-(f(x_i^a)-f(x_i^n))^2}{\beta}+1+\epsilon)],
\end{aligned}
\end{equation}

where $\beta$ is a scaling factor (which is equal to the dimensionality of the embedding space) and $\epsilon$ is a small value to prevent $\ln(0)$. This triplet loss function allows the network to learn even when the difference between samples is negative and converge faster.

\subsection{Training the Triplet-based model}

The main idea of a triplet loss model is to map the image into a high dimensional feature space, where ideally the embeddings representing the same identity will be located closer to each other than to those belonging to different identities. Using such space allows us to use a distance metric and assess whether the samples originate from the same person. We used the \textit{Combined} dataset for training and evaluating TLPIM. A random disjoint 80\%/20\% split was created for training and testing, respectively. Hard triplet mining \cite{Schroff_2015_CVPR,hermans2017defense} was used in the formation of triplets for training. 

We trained TLPIM with an initial learning rate of 0.00001 and \textit{Adam} optimizer, in batches of 24 triplets for a minimum of 20,000 iterations, interrupting the training if validation loss did not improve for 20 iterations. To perform recognition, we first extract the image embeddings with TLPIM, and then calculate the Euclidean distance between their 128-dimension coordinates. Using the distribution of distances for genuine and impostor pairs, it is possible to define a decision threshold to produce \textit{match} or \textit{non-match decisions}.

\section{Visual Explanations}

Despite its acceptance as a reliable means of identification, automated iris recognition methods still offer limited and non-standard methods of visualization to let human examiners interpret the model output. This is mainly due to pending research debate what are the best iris features that could be named and understood by human examiners, having at the same time adequate discriminatory strength. The goal of this work is to offer visualizations that can provide human experts hints regarding the image properties that led to an automatic matching decision. In addition to the enhanced individual segmentation masks generated by Mask R-CNN, we also applied Class Activation Maps (CAM) \cite{Zhou_2016_CVPR} for that purpose.

While more modern derivations of the original CAM may seem to offer attractive results, they are not appropriate for our use case. Grad-CAM \cite{Selvaraju_2017_ICCV} attempts to merge the coarse localization provided by CAM to the fine-grained feature activations yielded through guided backpropagation \cite{springenberg2015}. The main problem with the Grad-CAM approach is that it is targeted at classification problems, and requires a target label to calculate the gradients for backpropagation. A more recent work \cite{Chen2020AdaptingGF} tries to address this by making adaptations to problems in the embedding space, however their solution is more appropriate for closed-set settings, since a target embedding is calculated by averaging multiple embeddings of the same class. Since our application cannot rely on multiple enrollments of the same identity, we decided to use the traditional CAM approach to provide the activation maps produced by the input.

\begin{figure}[htb]
\begin{center}
   \includegraphics[width=1.0\linewidth,trim={12 0 12 0}]{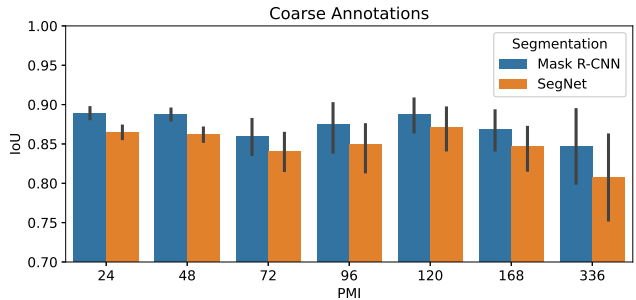}
\end{center}
   \caption{Segmentation accuracy by postmortem interval.}
\label{fig:segmpmi}
\end{figure}

\section{Results}

\subsection{Postmortem iris segmentation}

\begin{table}[]
\begin{center}
\begin{tabular}{lcc}
\hline
       & \multicolumn{2}{c}{Average IoU (\%)} \\ 
Annotations         & Mask R-CNN & SegNet\cite{troki_wacv2020} \\ \hline \hline
Coarse & 88.38 $\pm$ 6.54   & 85.97 $\pm$ 7.26      \\
Fine   & 86.82 $\pm$ 7.07   & 83.31 $\pm$ 7.62      \\
\hline
\end{tabular}
\end{center}
\caption{Segmentation accuracy in comparison to SegNet.}
\label{tab:segmentation}
\end{table}

\begin{figure*}[htb]
\begin{center}
   \includegraphics[width=.95\linewidth,trim={0 0 0 0}]{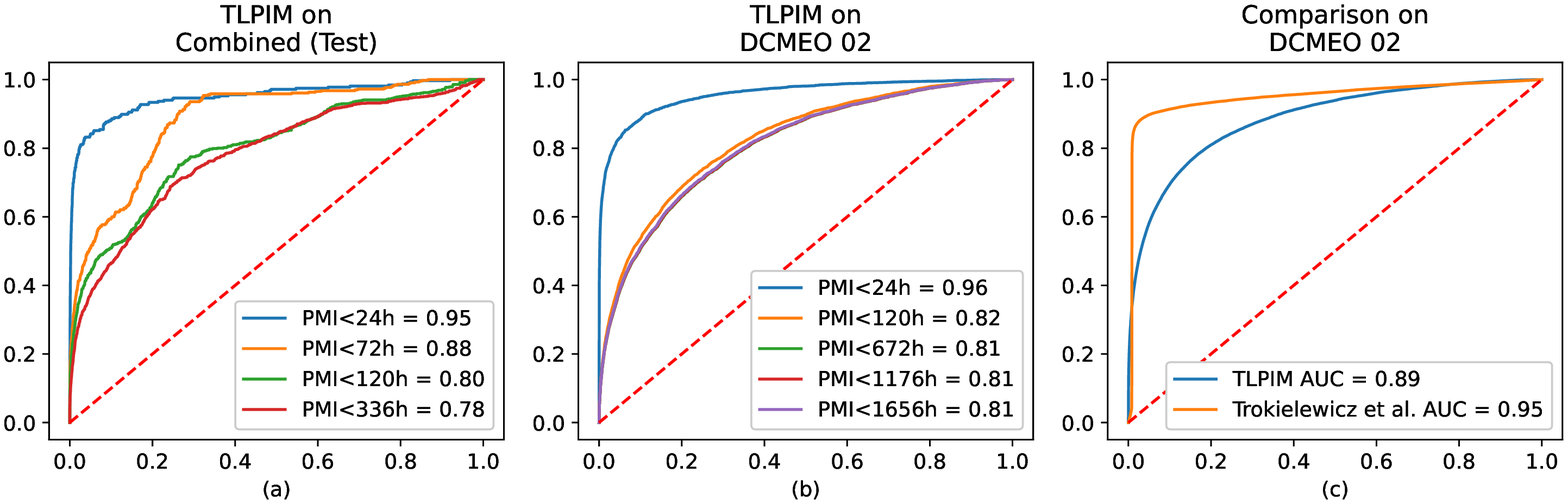}
\end{center}
   \caption{Matching results: (a) TLPIM performance by PMI on the {\it Combined} dataset; (b) TLPIM performance by PMI on the {\it DCMEO 02} dataset; (c) performance comparison between our approach and Trokielewicz \etal~\cite{troki_wacv2020}.
   }
\label{fig:roc_curves}
\end{figure*}

After training Mask R-CNN to perform iris segmentation using the segmentation dataset, we evaluate both models on the \textit{Combined} dataset. Using Intersection Over Union (IoU) to measure how well the predicted segmentation masks fit to the ground truth annotation, we compare our method to state of the art SegNet-based \cite{troki_wacv2020}. Since the annotations used to train SegNet provide two versions, \textit{Coarse} and \textit{Fine}, we trained and evaluated our model using both of them. Table \ref{tab:segmentation} shows our model achieves better accuracy and smaller variation than SegNet-based using both types of annotation. Given that the training and testing data are distinct, these results also indicate generalization capabilities.

An evaluation that takes into consideration the PMI was also conducted. Fig. \ref{fig:segmpmi} illustrates how the decay process constrains the accuracy of iris segmentation. Although, as the PMI increases the segmentation accuracy goes down and variance increases, the proposed Mask R-CNN-based segmenter consistently outperforms the previous one based on SegNet. Following these results, we adopted the coarse segmentation-based model for the remainder of this work.

We fine-tuned our Mask R-CNN model to perform simultaneous detection of specular highlights and wrinkles (Section \ref{sec:segmentation}). Finally, we combined the segmentation masks with the network class activation maps to produce a comprehensive visual explanation of the matching process (presented in detail in Sec. \ref{subsec:Visualization}). 

\subsection{Recognition}

After training on one partition of the \textit{Combined} dataset, we evaluated TLPIM on the remaining, subject-disjoint test partition of the same dataset, as well as on the newly collected \textit{DCMEO 02}. As a general performance metric, we calculated the Area Under the ROC (AUROC) for these datasets, on which TLPIM achieved 0.84 and 0.89 respectively.

We also evaluated the proposed solution as a function of the PMI range of the reference and probe samples (Fig. \ref{fig:roc_curves} (a) and (b)). Here, we could specifically measure the extent to which postmortem decay affects recognition. AUROC was as high as 0.95 and 0.96 on \textit{Combined} set and \textit{DCMEO 02} respectively, if we only consider pairs where PMI of both samples is under 24 hours. If we keep the same restriction on the reference samples ($<=$24h) and allow probe samples up until longer periods, the result is that accuracy drops proportionally to the increase in the PMI period. On the \textit{Combined} set, the AUROC is 0.88, 0.80 and 0.78 as we allow probe samples to be up to 72, 120 and 336 hours, respectively. Similarly, on the larger \textit{DCMEO 02}, the AUROC declines to 0.81 for probe samples with the PMI above 672 hours.

A comparison of our approach with the only existing postmortem end-to-end iris recognition method is presented in Fig \ref{fig:roc_curves}(c). The AUROC-wise performance of our method is slightly inferior to Trokielewicz \etal~\cite{troki_wacv2020}. The important distinction to be made here is that the proposed method offers interpretability through visualization, apparently at some small cost of accuracy.

\subsection{Visualization}
\label{subsec:Visualization} 

Interpretability and visualization are not frequent components in iris recognition systems. However, especially when machine learning methods are involved, there are scenarios where a human examiner would be still required to inspect and interpret the machine's results. With this goal, we created a visualization that allows examiners to understand multiple outputs of our method, as well as validate its output, in order to ground their expert decision.

Figure \ref{fig:pair_viz} shows visualizations for sample matching pairs. Each output produced by our model is represented as a layer in these images, which can be switched on and off independently, \eg to make the output less cluttered or focus on a particular artifact. The background is the contrast-enhanced input image, on top of which the contours of segmented objects are drawn -- while blue denotes the iris boundaries, yellow and green represent highlight and wrinkle occlusions, respectively. Lastly, an overlay with the CAM heatmap hints to the examiner the regions with higher importance in the extraction of the iris image features.

It is important to note that there are no (known to us) iris segmentation benchmarks that would allow for independent evaluation of segmentation accuracy of postmortem-specific iris artefacts, so quantitative evaluation of the resulting visualization methodology could not be performed.

By inspecting the matching pair visualization, the examiner can quickly assess whether the similarity score between the images is appropriate or it was influenced by mistakes at one or more stages of iris image processing. For instance, Fig \ref{fig:pair_viz} (b) and \ref{fig:pair_viz} (c) are cases where the examiner may decide the similarity scores should not be trusted. Despite both images have yielded good segmentation results and some minor degradation, the heatmaps reveal that for both pairs, significant areas outside the iris annulus played a major role in calculating the network embedding. In turn, pairs shown in (a) and (d) seem to be matched by a network using areas inside the iris annulus, what may suggest a stronger set of features being used in matching.

\subsection{Interpretation of visual outputs}

\begin{figure*}[htb]
\begin{center}
  \includegraphics[width=1.0\linewidth,trim={0 0 0 0}]{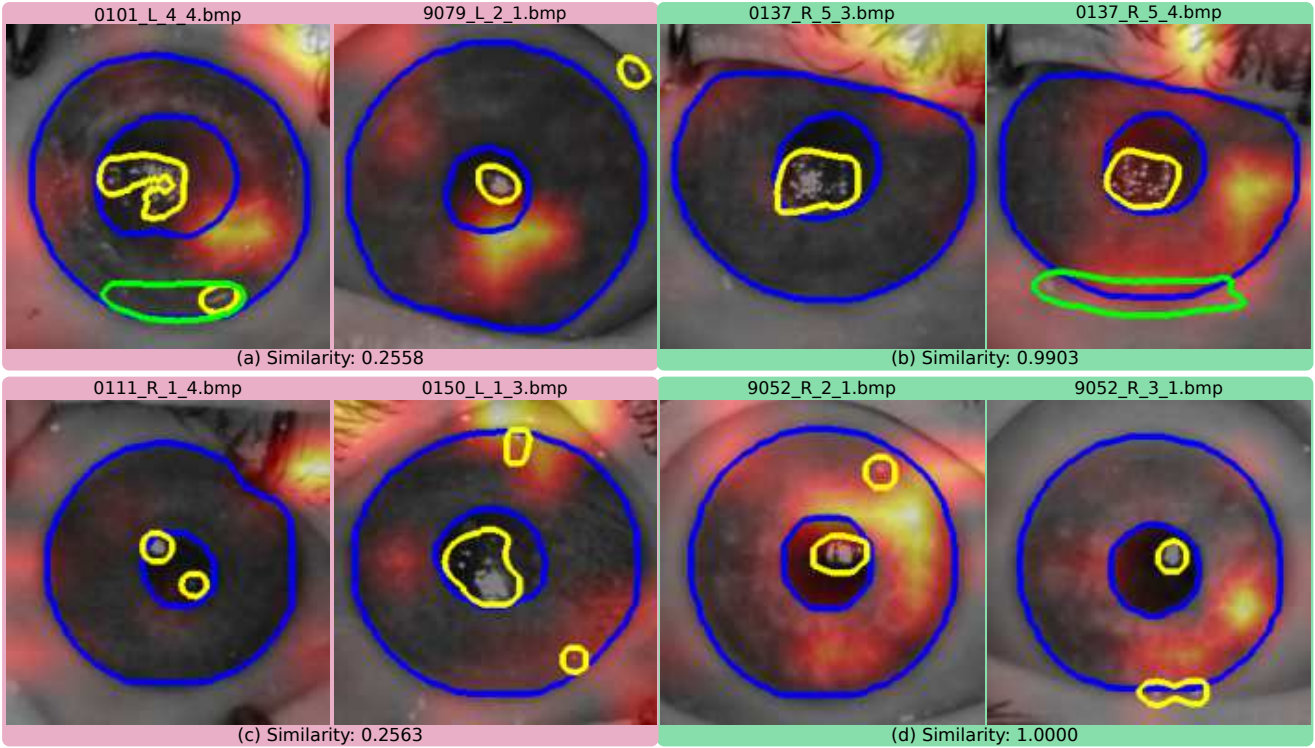}
  \caption{Visual inspection of matching pairs (genuine pairs marked in green and impostor pairs marked in red) is intended to provide an aid to experts making a final judgment on the analyzed pairs. Composite visualization aggregates three layers of information: 1) enhanced input image, 2) iris and postmortem artifacts boundaries, and 3) network activation heatmap.}
\label{fig:pair_viz}
\end{center}
\end{figure*}

While recognition accuracy is the most desirable quality in a biometric matching system, it is not the only goal of this work. Many recognition methods, especially the deep-learning based ones, yield very high accuracy rates, but offer essentially nothing that explains its output. Although the proposed method does not surpass the current state of the art in postmortem iris recognition, it delivers a substantial contribution to the interpretability of its decisions. By analyzing the visual output of TLPIM, forensic examiners can get insights regarding the quality of the algorithm quantitative and categorical output. By examining the results depicted in Figure \ref{fig:pair_viz} we offer examples of how visual output interpretation can offer valuable insight on the quality of the decisions.

\emph{Figure \ref{fig:pair_viz}(a)} -- This impostor pair resulted in low similarity score (0.2558), and consequently a non-match by the algorithm (denoted in the figure by the red box). Segmentation of the iris are adequate as well as the specular highlight and wrinkled occlusion areas in both images. A significant network activation occurs within the usable iris area, in regions approximately equivalent in both images. The strong activation located in the upper right area of the first image can be considered a demerit: it is outside the bounds of the iris annulus, and corresponds to the eyelash region and is therefore weak in identifying features.

\emph{Figure \ref{fig:pair_viz}(b)} -- A genuine pair, with good iris segmentation despite the substantial eyelid occlusion, and correct segmentation of the highlight constellations overlapping mostly the pupil regions. In the second image of the pair, there is a questionable identification of a wrinkled region at the bottom, but mostly outside the iris boundary. TLPIM declared this pair a match, as indicated by the green box around it. However, the strong similarity score for this pair (0.9903) can be disputed if one considers the potent activations in the eyelash regions of both images: they suggest the score relies more in the evident similarities between these areas, and not the irises.

\emph{Figure \ref{fig:pair_viz}(c)} -- This impostor pair was correctly classified as a non-match by TLPIM with a similarity score of 0.2563. Segmentation of the iris and highlights presents no obvious inaccuracies, and no wrinkled occlusion area was detected. Little could be challenged in this case, except for the fact that again the strongest network activations are in regions outside the iris boundaries.

\emph{Figure \ref{fig:pair_viz}(d)} -- This is perhaps the most indisputable decision out of these examples, this pair was deemed a match with the highest possible similarity score (1.0). In both images, it is possible to note very good segmentation, where only minor occlusions occur to the iris area. At the same time, one can observe the highest network activation areas are inside the iris boundaries, in regions that partially overlap between both images. In addition to the strong similarity score, visual analysis corroborates this pair as a highly probable match.

\section{Conclusions}

In this work, we have presented an interpretable end-to-end deep learning-based iris recognition system with a primary goal of providing interpretations to forensic and medical examiners. Typical scenarios for these professionals consist in comparing live iris images with those collected after death, containing eye decay-sourced artifacts. The existing iris recognition tools have shown to perform sub-optimally in postmortem scenarios, and did not offer visual explanations specific to postmortem iris recognition. This paper makes a first step in this direction. 

Using the visual explanations produced by the proposed TLPIM method, human experts can verify the results of iris matching scores, and in some sense assess the trustworthiness of the matcher. This is especially important when decay artifacts overlap with the iris annulus and may impact the comparison score. Furthermore, it is possible to easily identify whether such regions played a relevant part in the extraction of embeddings by the network, owing to the class activation mapping embedded into this tool.

Objectively, we developed the first (to the best of our knowledge) fully deep learning-based processing pipeline completely based on open source frameworks that is able to perform postmortem iris segmentation and recognition and provide human-interpretable insights into its decision process. We make the source code of the entire method available along with this paper\footnote{Available at \url{https://github.com/akuehlka/xai4b_tlpim.git}}. Additionally, the curated test dataset collected during this work will be submitted to the NACJD archives, and can be requested to facilitate further research in postmortem iris identification. 

\section{Acknowledgements}

This project was supported by Award No. 2018-DU-BX-0215, awarded by the National Institute of Justice, Office of Justice Programs,
 U.S. Department of Justice. The opinions, findings, and conclusions or recommendations expressed in this presentation are those of the authors and do not necessarily reflect those of the Department of Justice.

{\small
\bibliographystyle{ieee_fullname}
\bibliography{main}
}

\end{document}